\documentclass[conference]{IEEEtran}
\IEEEoverridecommandlockouts
\usepackage{cite}
\usepackage{amsmath,amssymb,amsfonts}
\usepackage{algorithmic}
\usepackage{graphicx}
\usepackage{textcomp}
\usepackage{float} 
\usepackage{stfloats}
\usepackage{subfigure} 
\usepackage{amsmath}
\usepackage{xcolor}
\def\BibTeX{{\rm B\kern-.05em{\sc i\kern-.025em b}\kern-.08em
    T\kern-.1667em\lower.7ex\hbox{E}\kern-.125emX}}
\begin{document}

\title{SRRM: Semantic Region Relation Model for Indoor Scene Recognition\\
\thanks{*Corresponding author.

This work was jointly supported by the Key Development Program for Basic Research of Shandong Province under Grant ZR2019ZD07, the National Natural Science Foundation of China-Regional Innovation Development Joint Fund Project under Grant U21A20486, the Fundamental Research Funds for the Central Universities under Grant 2022JC011.}
}

\author{\IEEEauthorblockN{Chuanxin Song}
\IEEEauthorblockA{\textit{School of Control Science and Engineering} \\
\textit{Shandong University}\\
Jinan, China \\
songchuanxin@mail.sdu.edu.cn}
\and
\IEEEauthorblockN{Xin Ma\textsuperscript{*}}
\IEEEauthorblockA{\textit{School of Control Science and Engineering} \\
\textit{Shandong University}\\
Jinan, China \\
maxin@sdu.edu.cn}
}

\maketitle

\begin{abstract}
Despite the remarkable success of convolutional neural networks in various computer vision tasks, recognizing indoor scenes still presents a significant challenge due to their complex composition. Consequently, effectively leveraging semantic information in the scene has been a key issue in advancing indoor scene recognition. Unfortunately, the accuracy of semantic segmentation has limited the effectiveness of existing approaches for leveraging semantic information. As a result, many of these approaches remain at the stage of auxiliary labeling or co-occurrence statistics, with few exploring the contextual relationships between the semantic elements directly within the scene. In this paper, we propose the Semantic Region Relationship Model (SRRM), which starts directly from the semantic information inside the scene. Specifically, SRRM adopts an adaptive and efficient approach to mitigate the negative impact of semantic ambiguity and then models the semantic region relationship to perform scene recognition. Additionally, to more comprehensively exploit the information contained in the scene, we combine the proposed SRRM with the PlacesCNN module to create the Combined Semantic Region Relation Model (CSRRM), and propose a novel information combining approach to effectively explore the complementary contents between them. CSRRM significantly outperforms the SOTA methods on the MIT Indoor 67, reduced Places365 dataset, and SUN RGB-D without retraining. The code is available at: https://github.com/ChuanxinSong/SRRM
\end{abstract}


\makeatletter
\renewcommand\footnoterule{%
  \kern-3\p@
  \hrule width \columnwidth
  \kern2.6\p@}
\makeatother

\section{Introduction}
Indoor scene recognition is a wide-ranging research topic in computer vision, which has been widely used in application fields such as smart cameras and intelligent robots. It is also considered as a prerequisite or prior knowledge for other computer vision tasks such as image retrieval and object detection, thanks to its ability to provide a basic description of the content of an image\cite{ref1}.

\begin{figure}[tbp]
    \centering
    \includegraphics[width=8cm]{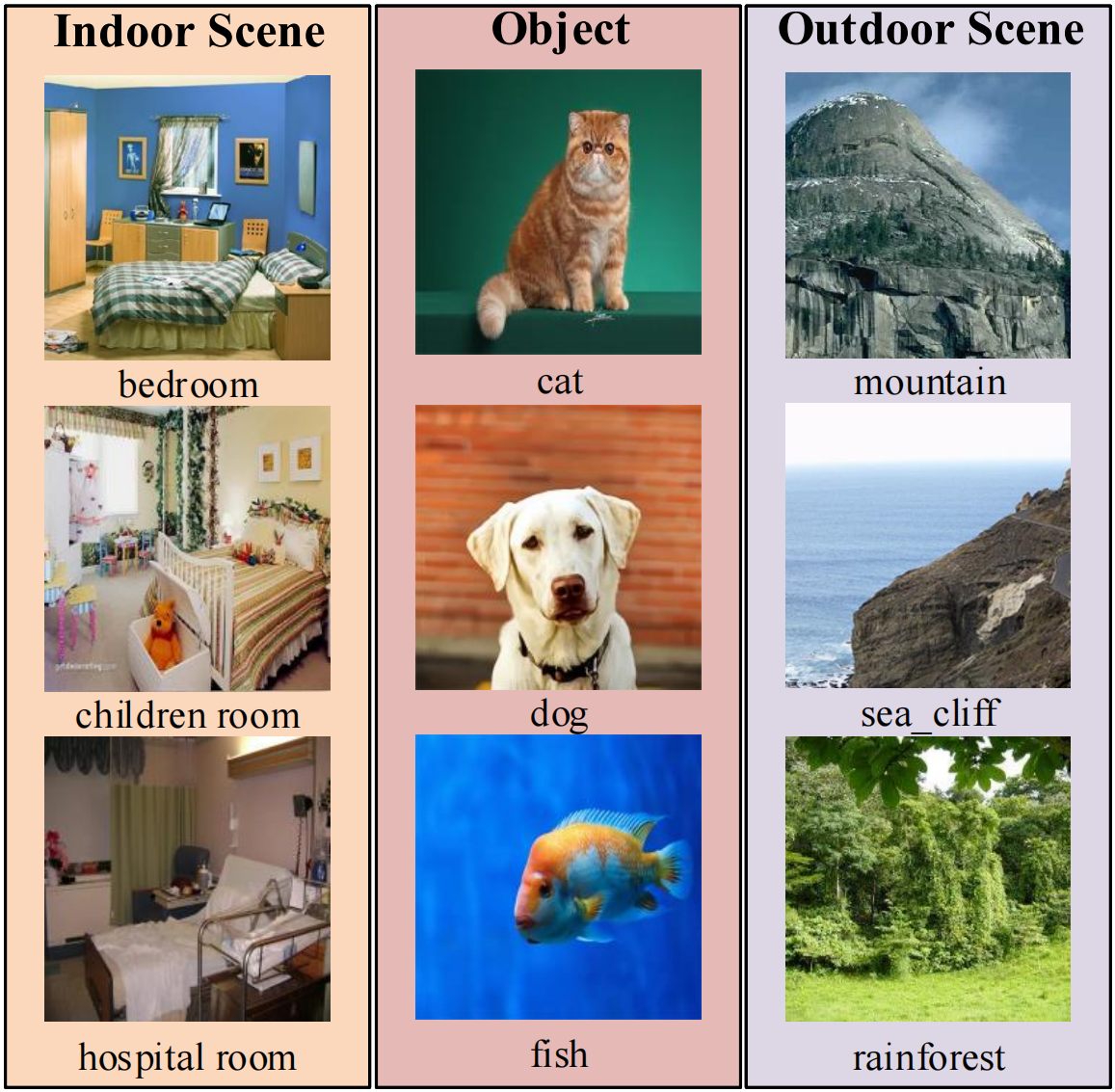}
    \caption{Some examples of different datasets (indoor scene, object, outdoor scene).}
    \label{Fig7}
\end{figure}

Deep neural networks (DNNs) have been successful in learning advanced representations of images, and they have been widely used in scene recognition tasks in recent years. However, as show in Fig. \ref{Fig7},  the indoor scenes, which contains multiple objects with complicated relationships,  are more incomprehensible than single-object or outdoor scenes. 
Moreover, the presence of co-existing objects from different scene classes often leads to a feature similarity phenomenon, which poses a challenge for DNNs to achieve comparable accuracy to object recognition or outdoor scene recognition in indoor scene recognition field. To overcome above limitations, we focus on exploring the contextual relationships of semantic information within scenes to better explain a given scene.

Similar strategies have been proposed recently\cite{ref7,ref18,ref23,ref17,ref19}, which obtain semantic information within scenes to assist scene recognition, with positive results. 

Specifically, approaches\cite{ref7,ref18,ref23} consider a statistical-like way to exclude non-discriminatory objects from the scene in terms of how often they appear in the scene. However, due to the complex and diverse indoor environment, the screening of discriminatory objects is quite difficult. There are also some approaches\cite{ref17,ref19} use semantic information to attach label meanings to features in the backbone network, constrain scene recognition by exploring the contextual relationships between feature regions with different semantic labels. However, the accuracy of the above methods mainly depends on the feature extraction capability of the backbone network, and the utilization of internal object information is limited. In addition, limited by the accuracy of semantic segmentation or object detection techniques, all methods combining semantic information for scene recognition inevitably face a problem, which is the negative impact of semantic ambiguity. To the best of our knowledge, existing methods use confidence thresholds to filter and mitigate semantic ambiguity, but the threshold method is obviously not flexible enough and has limited effect in the face of large data volume.

In this paper, we propose the semantic region relation model (SRRM), which differs from context-based methods that assign semantic labels to backbone features. Instead, we  aim to leverage the spatial relationships among semantic objects in a scene by exploring the semantic segmentation score tensor directly. This approach offers improved interpretability while also fully utilizing semantic information. Moreover, facing the negative effects of semantic ambiguity, the proposed method does not limit itself to rigidly dealing with the problem using confidence thresholds as its predecessors, but proposes a novel way to filter the ambiguity points adaptively according to the state of the image itself, yielding remarkable results.

Meanwhile, since SRRM which takes the semantic segmentation map as input does not consider some information such as color and texture in the image, we combine it with PlacesCNN which takes RGB pictures as input and generate global representations for them respectively, so as to deeply study the complementary information between the two branches.

In summary, our main contributions of this paper are as follows:
\begin{itemize}
\item We propose a novel framework SRRM that enables modeling semantic region relation directly on semantic segmentation results for indoor scene representation, and SRRM outperforms all existing methods that use only semantic segmentation results for scene recognition.
\item We propose an easy but efficient confidence filtering method to reduce the negative impact of semantic ambiguity. This approach can adaptively and significantly enhance the reliability of semantic segmentation results.
\item Meanwhile, in order to fully explore the information contained in the image, we combine the proposed SRRM and the PlacesCNN model as CSRRM, and propose a novel aggregation method to combine the output of the two modules, which better explore the complementary information between them.
\item We evaluate the effectiveness of the proposed method on MIT-67, Places365-7 and Places365-14 datasets, as well as its generalization performance on the SUN RGBD dataset, all of which yielded state-of-the-art results. 
\end{itemize}

\section{Related Works}

In this section, we review the research work related to scene recognition and discuss the differences and connections between these related works and our approach.

Scene recognition is an important research topic in computer vision. Many early approaches used local visual descriptors (such as LBP\cite{ref2}, SIFT\cite{ref3}, OTC\cite{ref4}, etc.), and use the BOVW framework\cite{ref6} to integrate these local visual descriptors into image representation. Quattoni et al.\cite{ref8} proposed a prototype based model for indoor scene that can combine local and global discriminative information. However, the features used by the above methods are all hand-crafted, which is limited to distinguish blurred or high similarity scenes.

In recent years, deep neural networks have made significant progress in computer vision tasks\cite{ref9,ref10,ref11}. Some approaches\cite{ref5,ref12,ref13,ref14} attempt to extract visual representations for scene recognition through convolutional neural networks. Dual CNN-DL\cite{ref13} proposed a new dictionary learning layer to replace the traditional FCL and ReLu, which simultaneously enhances the sparse representation and discriminative ability of features by determining the optimal dictionary. Lin et al.\cite{ref15} proposed to transform convolutional features to the ultimate image representation for scene recognition by a hierarchical coding algorithm. These methods use convolutional neural networks to extract scene representations, which greatly improve recognition results, but still fall far short of those achieved in tasks such as image classification and object detection, which stem from the fact that unrelated scene classes may share the same objects\cite{ref16}, and CNNs lack an effective representation of co-occurring objects within a scene. With this in mind, some methods combine the context of the objects in the scene to recognize the scene. 

Based on the context information, In SAS-Net\cite{ref17}, the semantic features generated by the semantic segmentation score tensor are used to add weights to different positions of the feature map generated by the RGB image, so that the network pays more attention to the discriminative regions in the scene image. DEDUCE\cite{ref18} obtained the binary feature vector corresponding to the object inside scene through the detection network, and then combined it with the backbone feature to assist the backbone network in indoor scene recognition. ARG-Net\cite{ref19} detects the foreground region and background region in the scene through semantic segmentation technology, and combines them with the feature map obtained by the backbone network to establish the context relationship (spatial relationship and morphological relationship) between regional features. The above methods try to establish the contextual relationship between the semantic regions in the scene to guide the scene classification. However, the object information in these methods only plays an auxiliary role, resulting in limited utilization of the internal object information.

In order to fully integrate the semantic object information in the scene, OTS-Net\cite{ref16} uses the semantic label graph to provide location information to the feature representation obtained by semantic segmentation down-sampling network, so as to perform scene recognition directly on the segmentation network through the self-attention mechanism, but ignores the diversity between the semantic segmentation task and the scene recognition task. Zhou et al.\cite{ref20} uses the probabilistic method to establish the co-occurrence relationship of objects in the scene, and combined the representative objects with the global representation of the scene to obtain a better scene representation, but the representation of object information inside the scene still stays on the surface. Meanwhile, to the best of our knowledge, the existing methods do not deeply consider the problem of detection error or segmentation error when using object detection or semantic segmentation technology for scene recognition.

In this work, we design the semantic region relation model, which fully investigates the relationship between semantic object regions inside scene, and addresses the ambiguity in semantic segmentation results through an adaptive approach.

\section{SEMANTIC REGION RELATION MODEL}

\begin{figure*}[htbp]
    \centering
    \includegraphics[width=15cm]{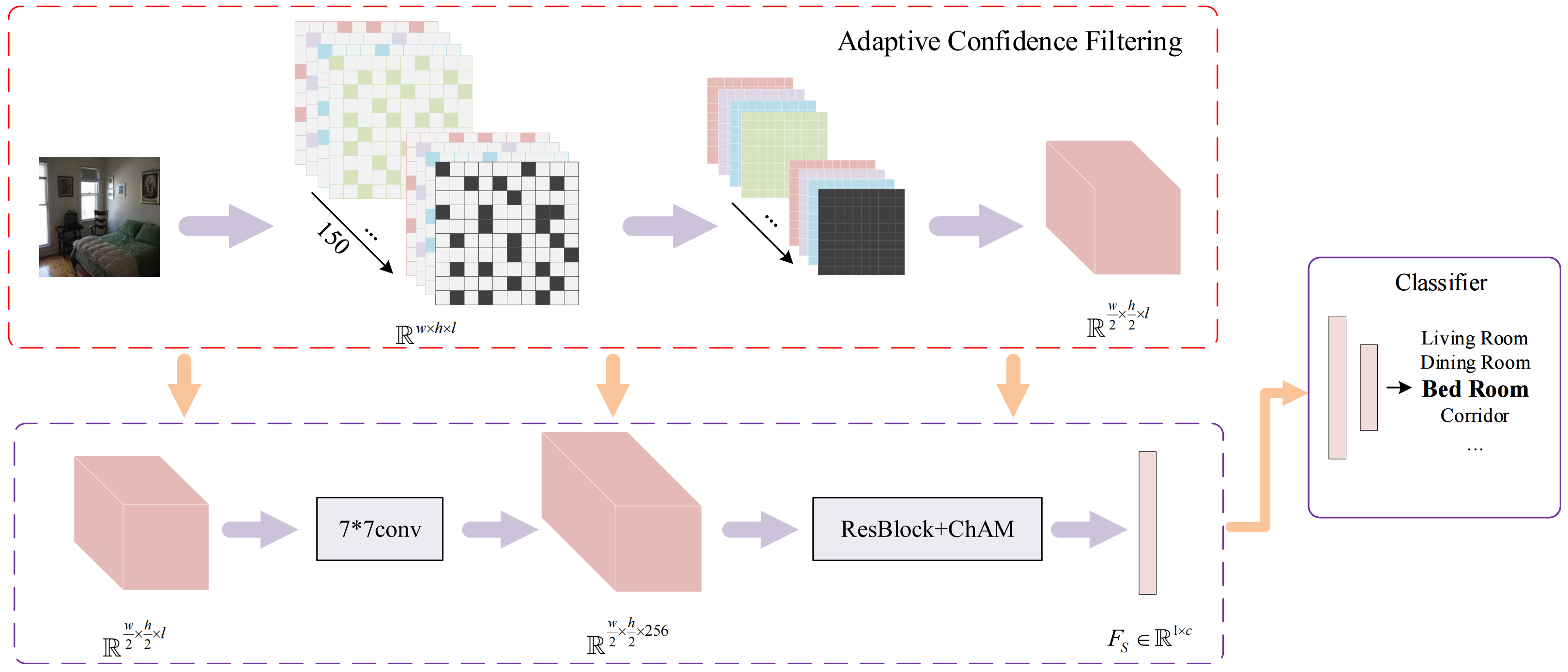}
    \caption{Semantic Region Relation Model(SRRM), where the part surrounded by the red dashed box represents the confidence filtering stage, is used to deal with the semantic segmentation error problem.}
    \label{Fig1}
\end{figure*}

We propose Semantic Region Relation Model (SRRM) for indoor scene recognition. In SRRM, The image $I \in \mathbb{R}^{w \times h \times 3}$ is first fed into the semantic segmentation network to generate the semantic segmentation score tensor $M \in \mathbb{R}^{w \times h \times l}$, which is then fed into the confidence filter. ${M_{i,j}} \in \mathbb{R}^{1 \times 1 \times l}$ represents the semantic prediction probability distribution of location $(i,j)$ in $I$, and there are $l$ semantic labels ($l=150$).  In this paper, Vision Transformer Adapter \cite{ref22} that is pretrained on ADE20K dataset\cite{ref23} is used as the segmentation network. The SRRM outputs a semantic global node features ${F_S} \in \mathbb{R}^{1 \times c}$, which is a high generalization of the semantic segmentation score tensor in both spatial and channel dimensions($c=2048$).

Based on the ResNet50 architecture, we designed a intuitive and clear network architecture for SRRM according to the specific channel characteristics of the segmentation score tensor $M \in \mathbb{R}^{w \times h \times l}$. Compared with the traditional ResNet50 architecture, the proposed network requires less computing power. Fig. \ref{Fig1} illustrates the framework of the proposed SRRM.

\begin{figure}[hbp]
    \centering
    \includegraphics[width=8cm]{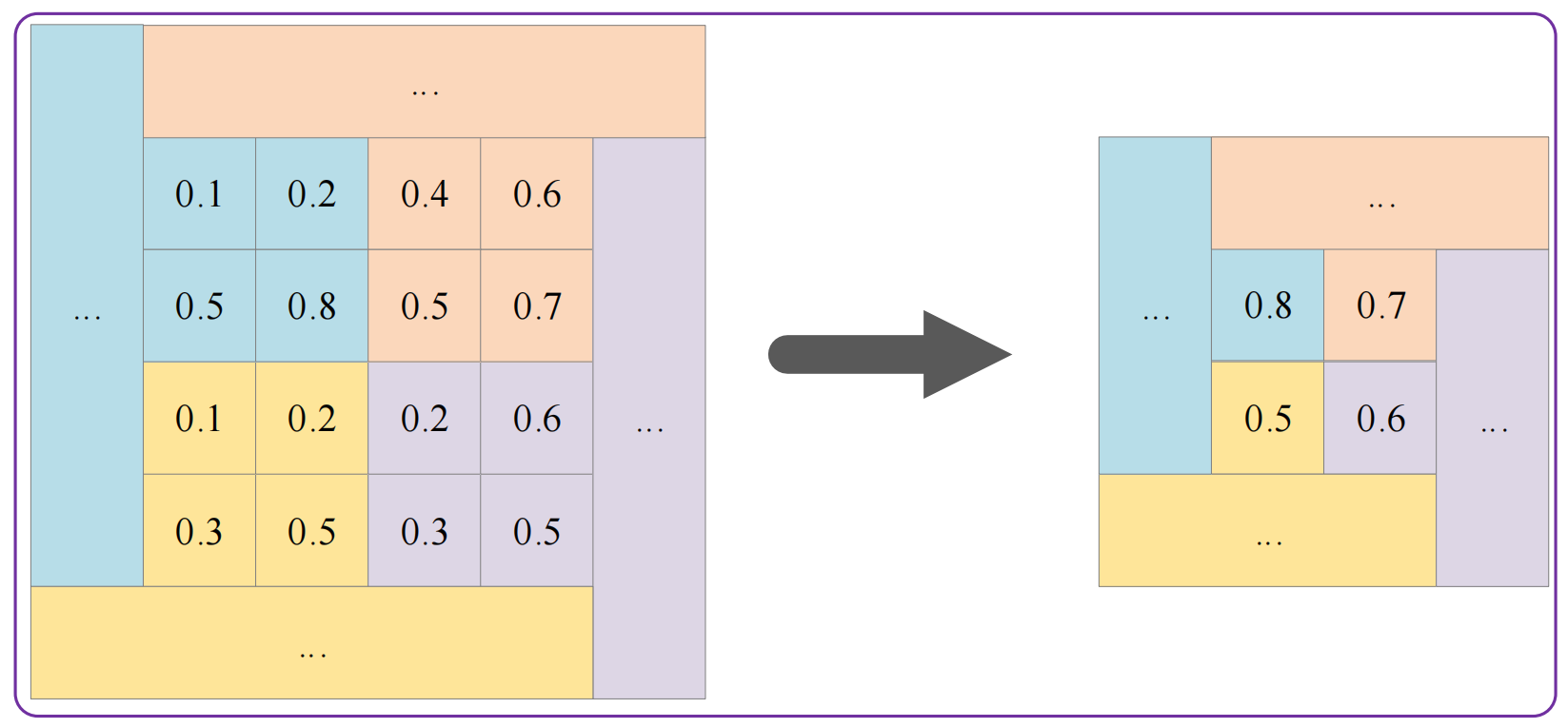}
    \caption{The Adaptive confidence filtering process in a single channel.}
    \label{Fig2}
\end{figure}

Due to the limited accuracy of semantic segmentation network, segmentation errors are inevitable. In order to reduce the negative impact caused by wrong semantic labels, we choose to first filter the semantic segmentation score tensor by confidence filter layer. In contrast to previous approaches, we do not rigidly rely on a certain threshold to filter semantic ambiguity points, but rather use an adaptive approach that allows for flexible filtering of each semantic segmentation score tensor. Specifically, We use a filter with a kernel size of $2 \times 2$ to smoothly process each channel of the score tensor, and for each filtered domain it passes through, the filter retains only the pixels with the highest confidence in its coverage, and outputs $M' \in \mathbb{R}^{\frac{w}{2} \times \frac{h}{2} \times l}$ after processing all channels of $M \in \mathbb{R}^{w \times h \times l}$. The confidence filtering process in a single channel is shown in Fig. \ref{Fig2}. Compared with the segmentation map corresponding to $M$, where the internal points represent the channels with the highest confidence in the $1 \times l$ range, each pixel point in the segmentation map corresponding to $M'$ represents the channel with the highest confidence in the $2 \times 2 \times l$ range in which it is located. In this way, our method filters the semantic segmentation graph using the coverage of the discriminative domain rather than a threshold, adaptively adjusting to each image's own state, improving both accuracy and high generalizability. Adaptive confidence filtering not only improves the reliability of the semantic segmentation score tensor, but also reduces the input size of the convolutional neural network, thereby reducing the computing power consumption of this module. We present a comparative justification of the idea in Section V.

Next, filtered semantic segmentation scores map $M' \in \mathbb{R}^{\frac{w}{2} \times \frac{h}{2} \times l}$ is processed using a convolutional neural network to extract features, Since the value of each channel in ${M'_{(i,j)}}$ represents the semantic prediction probability value of the pixel in $I$, inspired by SAS-Net\cite{ref17} and CBAM\cite{ref21}, the Channel Attention Module (ChAM)\cite{ref21} is introduced between convolutional blocks($7 \times 7$ conv not included), which can explore the relationship between different channels in the feature map. Since the channel values in the input score tensor represent the probabilities of the respective semantic categories, ChAM helps the network to better focus on the key semantic categories in the image by the probabilistic relationship. Specifically, given the feature map $F \in \mathbb{R}^{1 \times 1 \times {l^0}}$, ChAM first uses average pooling and Max pooling operations to aggregate the spatial information of $F$ to obtain feature vectors $F_{avg}$ and $F_{max}$. These vectors are processed by a shared multi-layer Perceptron (MLP) and then summed element-wise. After sigmoid activation, the channel attention map  $M_c \in \mathbb{R}^{1 \times 1 \times l^0}$ is obtained:

\begin{equation}
    \begin{aligned}
    {M_c}(F)&=\sigma (MLP({F_{avg}}) + MLP({F_{max}}))\\
    &=\sigma ({W_1}({W_0}Avg(F) + {W_1}({W_0}Max(F))
    \end{aligned}
    \label{eq1}
\end{equation}
where $\sigma$ denotes the sigmoid activation function, $W_0 \in \mathbb{R}^{{\frac{l^0}{r}\times l^0 }}$ and $W_1 \in \mathbb{R}^{{l^0 \times \frac{l^0}{r}}}$ are the weights of the MLP, and the ReLU activation function is followed by $W_0$, $r=16$ is the reduction ratio.

\begin{figure}[htbp]
    \centering
    \includegraphics[width=8cm]{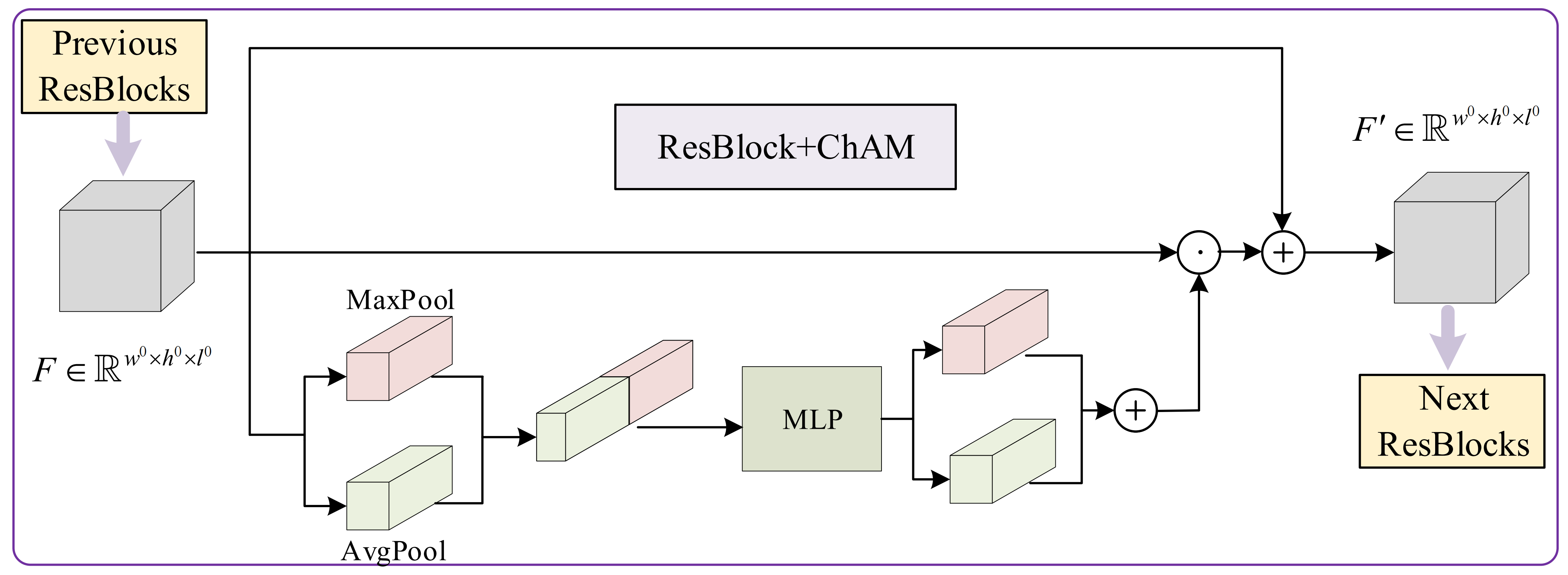}
    \caption{ResBlock + ChAM.}
    \label{Fig3}
\end{figure}

After $M' \in \mathbb{R}^{\frac{w}{2} \times \frac{h}{2} \times l}$ is processed by $7 \times 7$ conv and ResBlock+ChAM, the semantic global feature $F_S \in \mathbb{R}^{1 \times c}$ can be obtained, ResBlock includes the original ResNet-50’s three Basic Blocks (Basic Block 2, 3 and 4), ResBlock with ChAM added is shown in Fig. \ref{Fig3}, the channel attention map $M_c(F)$ is used to weight $F$ by:
\begin{equation}
    F' = F \oplus {M_c}(F) \odot F
    \label{eq2}
\end{equation}
where $\odot $ represents a Hadamard product, $\oplus$ represents Element-wise addition.

\section{CSRRM MODEL}

\begin{figure}[htbp]
    \centering
    \includegraphics[width=8cm]{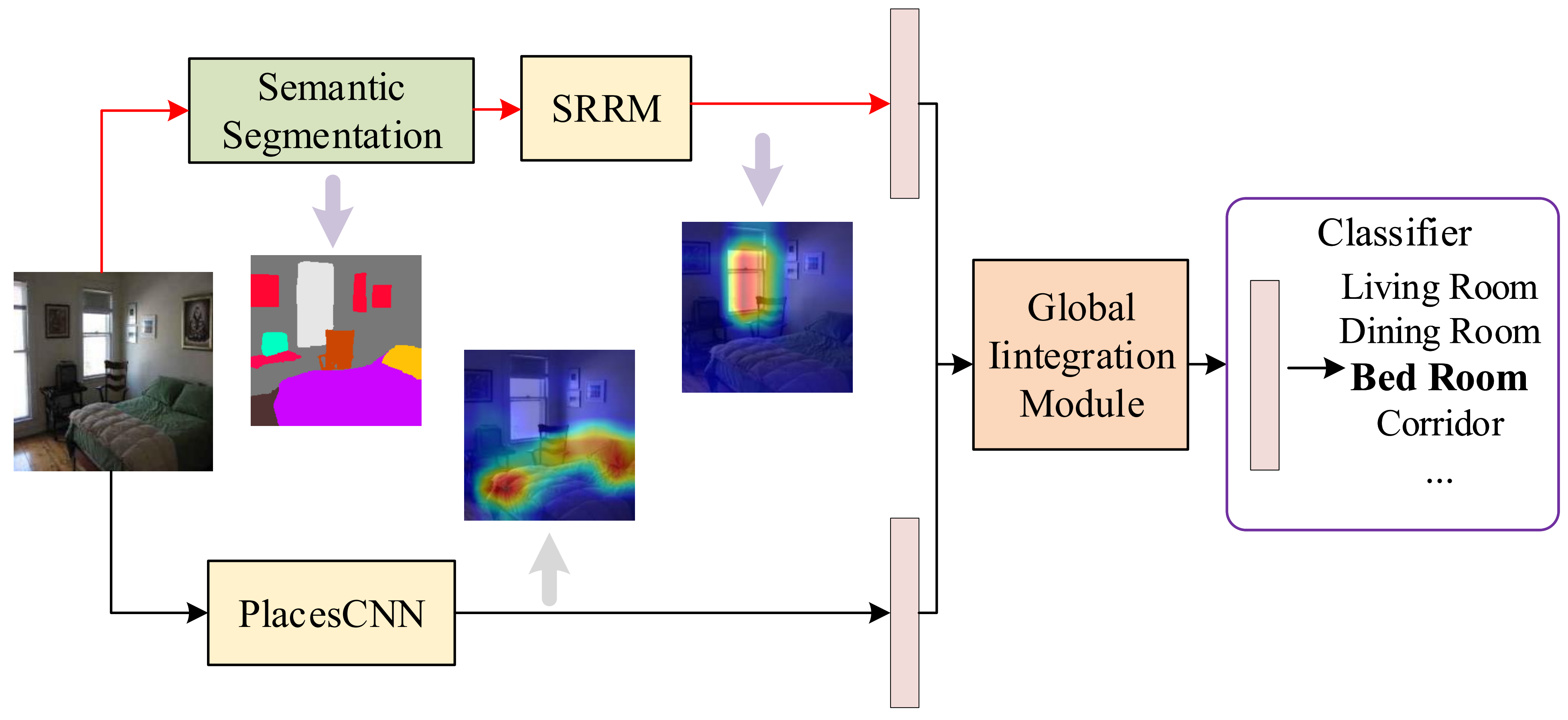}
    \caption{The proposed Combined Semantic Region Relation Model (CSRRM) contains two streams. The stream with the red arrow is the proposed SRRM that uses the semantic segmentation score tensor for scene recognition, the other stream with the black arrow is the PlacesCNN module that uses the raw RGB image for scene recognition.}
    \label{Fig4}
\end{figure}

Considering that the semantic segmentation score tensor input to the SRRM lacks information such as color and texture of the input image, in order to fully explore the information contained in the image, we combine the proposed SRRM and the PlacesCNN model as CSRRM, Fig. \ref{Fig4} show the overall architecture.

\subsection{PlacesCNN Module}

In this module, PlacesCNN model [14] with the base architecture ResNet50\cite{ref14}(without Pooling Layer) is used as the backbone network. The input of this module is the original RGB image $I \in \mathbb{R}^{w \times h \times 3}$, and the output is a global node feature $F_R \in \mathbb{R}^{1 \times c}$, where $c=2048$. For fair comparison, the module is pretrained on Places365 dataset\cite{ref14} when evaluate on MIT67 dataset\cite{ref8}, when evaluated on reduced Places365 and SUNRGB-D\cite{ref30} dataset, the module is pretrained on ImageNet dataset\cite{ref32}.

\subsection{Global Integration Module}

\begin{figure*}[htbp]
    \centering
    \includegraphics[width=16cm]{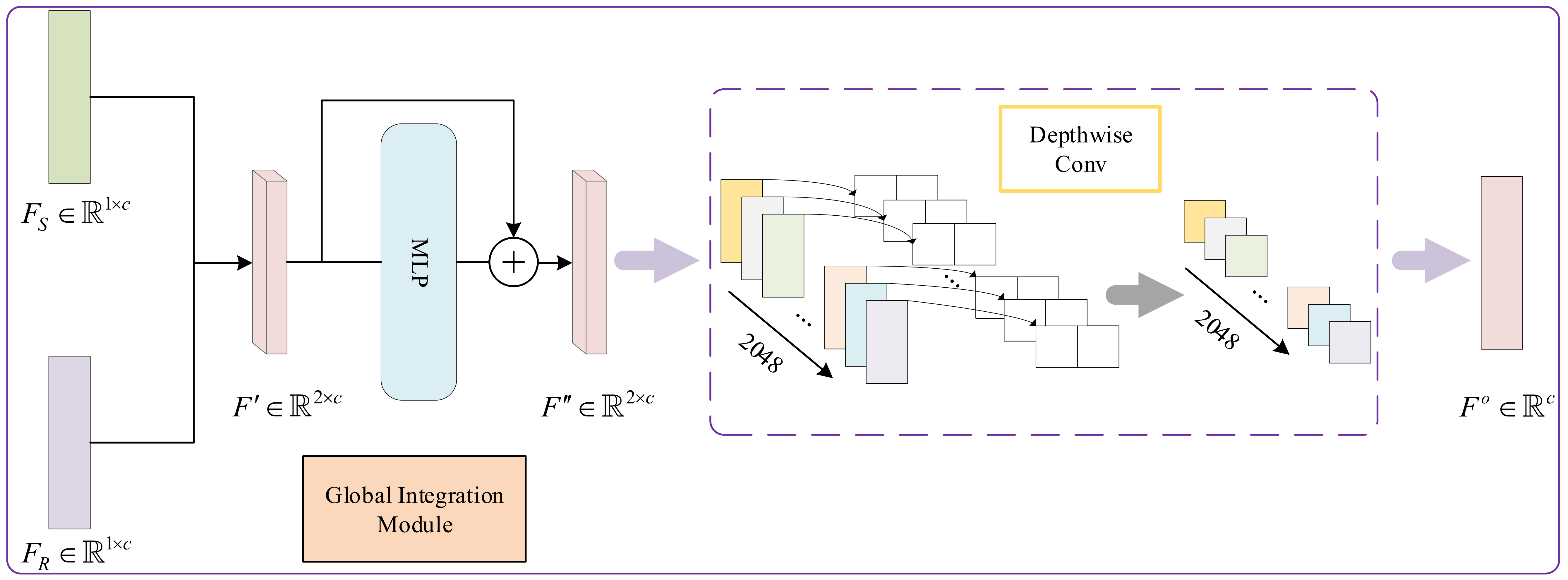}
    \caption{Global Integration Module, where the part surrounded by dotted boxes represents Depth wise convolution, which studies the complementary information of two global nodes from the channel dimension}
    \label{Fig5}
\end{figure*}

The Global Integration Module is used to explore the complementary information between the global node features $F_R \in \mathbb{R}^{1 \times c}$ and $F_S \in \mathbb{R}^{1 \times c}$, and obtain the aggregated feature $F^o \in \mathbb{R}^c$. Mining complementary information of two global features is key to accurate scene recognition, and inspired by MobileNet\cite{ref25} and OTS-Net\cite{ref16}, we propose a novel way to aggregate two global vectors from a channel perspective, namely, strip Depth-wise convolution, as shown in Fig. \ref{Fig5}. In section V, we will compare it with other aggregation architectures and demonstrate its superiority. The algorithm is introduced in the following.

Concatenating $F_R$ and $F_S$ in each channel dimension to obtain $F' \in \mathbb{R}^{2 \times c}$, which is then processed using a Multi-Layer Perceptron (using a residual structure to avoid overfitting) and outputs $F'' \in \mathbb{R}^{2 \times c}$:
\begin{equation}
    F'' = F' + \phi ({W_n}(Drop(\phi ({W_m}F' + {b_m}))) + {b_n})
    \label{eq3}
\end{equation}
where $\phi$  denotes the GeLu activation function, $W_m \in \mathbb{R}^{l \times c}$, $b_m \in \mathbb{R}^l$ and $W_n \in \mathbb{R}^{l \times c}$, $b_n \in \mathbb{R}^{c}$ are the weights and biases of the two FC layers inside the MLP, where $l=8192$. $Drop$ represents omit regularization ($dropout$) with a rate pf 0.1.

After MLP modification, $F''$ is input into DW (Depth-Wise) convolutional layer, where the two global node features are integrated in channel dimension. After deeply exploring the complementary information between them, the global integration feature $F^o \in \mathbb{R}^c$ can be obtained, and each channel value of $F^o$ can be expressed as:

\begin{equation}
    F_{m}^{o}=\sum_{i=1}^{2}{K_{i,m}\cdot F_{i,m}^{''} }  
    \label{eq4}
\end{equation}
where $K$ is the depth convolution kernel of size $2 \times 1 \times M$ where the $m_{th}$ filter in $K$ is applied to the $m_{th}$ channel in $F^{''}$ to produce the $m_{th}$ channel of $F^o$, $M=2048$.

Global aggregation feature $F^o$ is fed into a FC classifier to obtain the final scene prediction.

\subsection{Training procedure}

Due to the differences in the focus of RGB features and semantic features, when training two branch networks at the same time, the RGB and semantic modification center of gravity of some images are different, and the overall loss will hinder the optimization of less discriminative feature modules. To prevent this from happening, a two-stage training procedure is used when training validation is performed on the CSRRM.

In the first stage, the PlacesCNN branch and the SRRM branch are trained separately. In the second stage, the weights of the two branches are frozen and the global integration module is trained from scratch again.

\section{EXPERIMENTS}

In this section, we evaluate the effectiveness of proposed method on publicly available indoor scene recognition datasets: MIT-67\cite{ref8}, the reduced Places365\cite{ref14} and the reduced SUNRGB-D\cite{ref30}. The following subsections first introduce the benchmark datasets and then conduct ablation experiments to determine how each module affects the proposed method. After that, we will make a comparison between the proposed method and the existing state of the art methods.

\subsection{Implementation Details}

We use the Adaptive Learning-rates for Interpolation with Gradients (ALI-G) \cite{ref31} algorithm to optimize the trainable parameters in the network. Ali-G is an optimization algorithm for deep learning. This optimizer produces slightly lower performance than the SGD optimizer, but only requires the initial learning rate hyperparameter and does not require hand-crafted learning rate decay plans, which lightens the training process. In all our experiments, the initial learning rate is set to 0.1. During the second stage of training, in order to prevent overfitting, the dropout regularization function was used in the final classifier with an omission probability of 0.8. When evaluating the performance of the proposed method, we adopt the standard 10-crop testing method\cite{ref33}.This widely used evaluation methodology involves randomly extracting ten crops from an image and classifying each crop independently. The final classification is obtained by taking the average of the probabilities of the ten crops.

\subsection{Datasets}

\textbf{MIT-67 Dataset }\cite{ref8} consists of 67 indoor scene classes with a total of 15620 images, and each scene category contains at least 100 images. Following the recommendations by the authors\cite{ref8}, each class has 80 images for training and 20 images for testing. Due to the large intra-class variation of indoor scenes, the evaluation of MIT dataset is challenging.

\textbf{Places365 Dataset} \cite{ref14} is the largest and most challenging scene classification dataset to date, containing a wide range of indoor and outdoor scene categories. In this paper, a simplified version of it is used and only indoor scene categories are considered. For a fair comparison with other state of the art indoor scene recognition methods [16][20], we used the same two scene class Settings as them, namely Places365-7 and Places365-14. Places365-7 contains seven indoor scenes: Bathroom, Bedroom, Corridor, Dining room, Kitchen, Living room, and Office. Places365-14 contains 14 indoor scenes: Balcony, Bedroom, Dining room, Home office, Kitchen, Living room, Staircase, Bathroom, Closet, Garage, Home theater, Laundromat, Playroom, and Wet bar. The setup of the test set is the same as the official dataset \cite{ref14}.

\textbf{SUN RGB-D Dataset} \cite{ref30} is currently the largest RGB-D dataset. It is compromised of 3874 Microsoft Kinect v2images, 3389 Asus Xtion images, 2003 Microsoft Kinect v1 images and 1159 Intel RealSense images. The diversity of categories and sources makes SUNRGBD more suitable for verifying the generalization ability of the algorithm, so we used the reduced SUN RGB-D Dataset of the same class as Places365-7, and we test our model pretrained on the Place365-7 on SUN RGB-D dataset without retraining.

\subsection{Ablation studies}

In this part, we conduct ablation studies to evaluate the effectiveness of the proposed method. First, the impact of different SRRM architectures is evaluated. Second, the impact of different ways of combining two global features is evaluated. Unless specifically mentioned, the datasets used in ablation experiments is MIT-67 datasets. 

\textbf{1) Influence of the Semantic Region Relation Model}

In this subsection, only the SRRM is used for scene recognition, and the influence of network structure changes (Adaptive filter, Channel Attention Module) on the recognition effect is studied. The experimental results are shown in Table \ref{tab1}.

\begin{table}[htbp]
    \centering
    \caption{Ablation results for different architectures for the SRRM}
    \begin{tabular}{ccc}
        \hline
        \textbf{Architecture} & \textbf{Accuracy} & \textbf{Flops(G)} \\ \hline
        resnet50 & 64.403 & 27.2  \\ 
        4 * 4 filtering + resnet50 & 70.149 & 9.31  \\ 
        2 * 2 filtering +resnet50 & 69.627 & 9.31  \\ 
        resnet50\_ChAM & 69.851 & 27.3  \\ 
        4 * 4 filtering + resnet50\_ChAM & 73.284 & 9.32  \\ 
        2 * 2 filtering +resnet50\_ChAM & 74.403 & 9.32  \\ 
        \# 2 * 2 filtering +resnet50\_ChAM & 81.642 & 9.32  \\ \hline
        \multicolumn{3}{l}{\# indicates that the model's parameters are pre-trained on Places365.}
    \end{tabular}
    \label{tab1}
\end{table}

Results from Table \ref{tab1} suggest that filtering the semantic segmentation score tensor with confidence results in a huge improvement in network performance. Compared with the original semantic segmentation score tensor, the recognition accuracy of the network can be improved by 3.43$\%$ to 5.75$\%$ and the Flops can be reduced by 17.89G by inputting the filtered score tensor. For a given input, the channel attention module can improve the recognition accuracy of the network by 3.1$\%$ to 5.5$\%$, while the Flops only increases by 0.1G.
Meanwhile, we explore the influence of different confidence filtering domains on the recognition accuracy of the network. It can be found that after adding the Channel Attention Module to the network, the recognition accuracy generated by using the 2*2 filter domain to process the score tensor is 1.12$\%$ higher than that generated by the 4*4 filter domain. This is because although the confidence filtering operation can reduce the negative impact of semantic segmentation error points, it may also cause the segmentation map to lose a certain amount of object information. Compared with the 2*2 size filtering field, the 4*4 size filtering field will make the semantic segmentation score tensor lose more object information, which indicates that the size of the confidence filtering field should be considered from two perspectives of object information loss and semantic segmentation error. However, even if more object information is lost, compared with inputting the original semantic segmentation score tensor to the network, the 4*4 size filtering domain still improves the recognition accuracy by 3.43$\%$, which also illustrates the necessity of filtering semantic segmentation error points.

According to the above experiments, in the SRRM, we choose to first process the semantic segmentation score tensor using a confidence filter domain of size $2 \times 2$, and then use the resnet50 architecture with the channel attention mechanism for subsequent processing. At the same time, as shown in the last line of Table \ref{tab1}, we pre-trained the finally selected model on the Places365 dataset, which made it achieve higher precision on MIT67.

\begin{figure}[htbp]
    \centering
    \includegraphics[width=8cm]{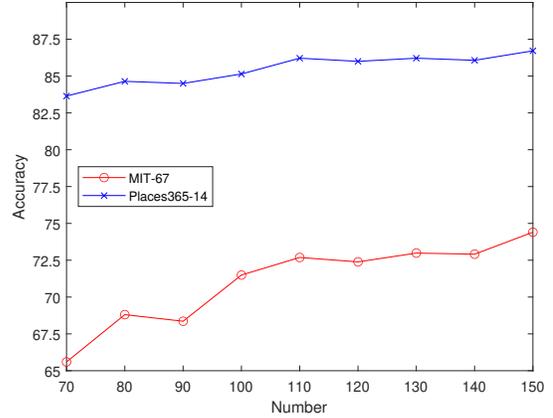}
    \caption{Ablation study of SRRM with different number of object knowledge ranging from 70 to 150 categories as shown in horizontal axis. The vertical axis shows the accuracy on percentage.}
    \label{Fig6}
\end{figure}

\textbf{2) Influence of the object knowledge}

To evaluate the impact of object knowledge on the SRRM, we conduct a set of ablation experiments on the MIT-67 and Places365-14 datasets, respectively. The result is shown in Fig. \ref{Fig6}. The horizontal axis represents the amount of object information used by the SRRM, and its value ranges from 70 to 150 (value interval is 10). We obtain the required object information in the order of the object list of ADE20K. Obviously, the accuracy of scene recognition is positively correlated with the amount of object information, and the recognition accuracy is greatly improved with the increase of the amount. For example, when using 150 semantic categories compared with using 70 semantic categories, the recognition accuracy of the former is 8.81$\%$ higher than that of the latter on the MIT-67 dataset, and the recognition accuracy of the former is 3.073$\%$ higher than that of the latter on the Places365-14 dataset. Meanwhile, we can find that with the increase of the number of object categories, the recognition accuracy does not always increase, and sometimes the accuracy changes little or even slightly. For example, when the number of semantic categories increases from 80 to 90, on the Places365-14 dataset, the recognition accuracy corresponding to 90 is 0.143$\%$ lower than that corresponding to 80. On the MIT-67 dataset, the recognition accuracy corresponding to 90 is also 0.448$\%$ lower than that corresponding to 80. This is because the added object categories at this time are cars, trucks, street lights, towers and other object categories unrelated to indoor scenes, so it seems that the correlation between semantic categories of object models and scene recognition is also crucial for scene recognition.

\textbf{3) The way two global node features are combined}

In this subsection, various ways are tried to aggregate the global node features to verify the effectiveness of the proposed method, which is also compared with two baselines. The experimental results are shown in Table \ref{tab2}, where Concatenation has a similar use in BORM\cite{ref20} and Semantic Gating Combination has a similar use in SAS-Net\cite{ref17}.

\begin{table}[htbp]
    \centering
    \caption{Ablation results for different architectures of global aggregation modules.}
    \begin{tabular}{ccc}
    \hline
        \textbf{Ways of combination} & \textbf{Accuracy} & \textbf{Flops(k)}  \\ \hline
        PlacesCNN(ResNet50) & 84.776 & -  \\ 
        SRRM & 81.642 & -  \\ 
        Depth-wise conv & 88.731 & 143.4  \\ 
        Concatenation & 88.582 & 274.4  \\ 
        Semantic Gating & 85.0 & 137.2  \\ \hline
    \end{tabular}
    \label{tab2}
\end{table}

Results from Table \ref{tab2} suggest that using either combination method yields better performance than the baseline method, which indicates that it is practical and effective to deeply explore the complementary information of semantic global features and RGB global features in the scene recognition task. Compared with the Concatenation, Depth-Wise conv not only consumes lower computing power resources, but also generates higher accuracy, which shows the superiority of DW convolution in exploring complementary information. Unexpectedly, the Semantic Gating Combination, which was outstanding in SAS-Net, produced the lowest recognition performance, improving only 0.224$\%$ over the baseline. This may be because the sigmoid activation function changes the scale of the semantic global features. It adversely affects its combination with RGB global features and is not suitable for our method.

In summary, we finally selected the best performing DW convolution aggregating two global node features.

\subsection{State-of-the-art comparison}

Along this section, the proposed approach is compared with the existing state-of-the-art methods. Comparison is performed on four indoor scene datasets: MIT-67\cite{ref8}, Places365-7\cite{ref14}, Places365-14\cite{ref14}, Reduced SUN RGB-D\cite{ref30}, when validating on reduced SUN RGB-D, the model are pre-trained on Place365-7 without retraining to compare the generalization performance. Unless explicitly mentioned, results of all the methods are extracted from their respective papers.

\begin{table}[htbp]
    \centering
    \caption{Comparison with the state-of-the-art methods by SRRM}
    \begin{tabular}{ccccc}
    \hline
        \textbf{Approaches} & \textbf{Places-14} & \textbf{Places-7} & \textbf{MIT-67 }& \textbf{SUN}  \\ \hline
        Deduce(${\Phi _{obj}}$)\cite{ref18}  & 47.0 & 62.6 & - & 53.6  \\ 
        BORM\cite{ref20}  & 74.9 & 83.1 & - & 69.2  \\ 
        OTS-Net\cite{ref16}  & 85.9 & 90.1 & - & 70.6  \\ 
        SAS-Net(Sem)[\cite{ref17}  & - & - & 73.43 & -  \\ 
        SRRM & 86.714 & 93.143 & 81.642 & 76.119  \\ \hline
    \end{tabular}
    \label{tab3}
\end{table}

\textbf{1) Comparison by SRRM}

In this section, SRRM is compared with the existing scene recognition methods that only use semantic representation obtained by semantic segmentation or object detection. The results are shown in Table \ref{tab3}. For fair comparison, SRRM is pre-trained on Places365 dataset for validation on MIT-67 and trained from scratch for validation on reduced Places365 dataset. The superior performance of our SRRM demonstrates that modeling semantic region relation is effective for scene recognition. 

\textbf{2) Comparison by CSRRM}

In this section, we compare the proposed CSRRM with existing state-of-the-art approaches on benchmark datasets, and the results are shown in Table \ref{tab4},\ref{tab5}, and \ref{tab6}, respectively.

\begin{table}[htbp]
    \centering
    \caption{State-of-the-art results on MIT-67 dataset.}
    \begin{tabular}{cc}
    \hline
        \textbf{Approaches}  & \textbf{Accuracy}  \\ \hline
        Recognition Indoor Scene\cite{ref8}   & 27  \\ 
        Places365+VGGNet16\cite{ref14}   & 76.53  \\ 
        Multi-Scale CNN\cite{ref12}   & 80.97  \\ 
        Dual CNN-DL\cite{ref13}   & 76.56  \\ 
        NNSD+ICLC\cite{ref15}   & 84.3  \\ 
        DAG-CNN\cite{ref26}   & 83.75  \\ 
        MP\cite{ref27}   & 86.9  \\ 
        SDO\cite{ref7}   & 86.76  \\ 
        SAS-Net\cite{ref17}   & 87.1  \\ 
        DeepScene-Net\cite{ref28}   & 71.0  \\ 
        ARG-Net\cite{ref19}   & 88.13  \\ 
        MR-Net\cite{ref24}   & 88.08  \\ 
        Ours  & \textbf{88.731}  \\ \hline
    \end{tabular}
    \label{tab4}
\end{table}

It is obvious that our approach substantially outperforms most existing indoor scene recognition methods in terms of effectiveness and generalization. Compared to the methods\cite{ref17,ref20,ref16,ref7,ref19,ref18} that also utilizes semantic information for scene recognition, our method achieves better effect, which demonstrates that it is very effective to explore the high-level representation of semantic information, and also proves the feasibility of in-depth study of complementary information between RGB and semantic representations. Moreover, our CSRRM also outperforms the current multi-branch-based approaches\cite{ref12,ref26,ref27,ref17,ref28,ref24,ref20} that obtain the multi-scale information of the scene,  which shows that it is effective to use semantic information as an additional source of information and obtain its high-level representation. Therefore, all these experiments have confirmed the superiority and generalization of the proposed method for indoor scene recognition.

\begin{table}[htbp]
    \centering
    \caption{State-of-the-art results on Places365-14 dataset.}
    \begin{tabular}{cc}
    \hline
        \textbf{Approaches}  & \textbf{Accuracy}  \\ \hline
        Word2Vec\cite{ref29}   & 83.7  \\ 
        BORM-Net\cite{ref20}   & 85.8  \\ 
        OTS-Net\cite{ref16}   & 85.9  \\ 
        Ours  & \textbf{88.714}  \\ \hline
    \end{tabular}
    \label{tab5}
\end{table}

\begin{table}[htbp]
    \centering
    \caption{State-of-the-art results on Places365-7 and reduced SUNRGB-D dataset.}
    \begin{tabular}{ccc}
    \hline
        \textbf{Approaches}  & \textbf{Places-7}  & \textbf{SUN RGB-D}  \\ \hline
        Deduce\cite{ref18}   & 88.1  & 70.1  \\ 
        BORM-Net\cite{ref20}   & 90.1  & 72.1  \\ 
        OTS-Net\cite{ref16}   & 90.1  & 70.6  \\ 
        Ours  & \textbf{93.429}  & \textbf{75.349 } \\ \hline
    \end{tabular}
    \label{tab6}
\end{table}

\section{Conclusions}

In this paper, we aim to develop a semantic region relation model for indoor  scene recognition. Initially, an adaptive confidence filtering module is proposed to mitigate the negative impact of errors in semantic segmentation. This module provides an innovative and effective idea for filtering semantic ambiguity points in approaches that utilize semantic segmentation. Based on this, we propose SRRM, which directly utilizes the semantic segmentation results to establish a global representation of the semantic region relations in the scene. 

Furthermore, given that the input of SRRM, i.e., the semantic segmentation score tensor, lacks detailed information such as color and texture, we integrate SRRM with PlacesCNN, which takes RGB as input. We design a novel global integration module to deeply explore the complementary information between  the two models.  The experiments demonstrate that our approach is more competitive compared to the existing methods.

In future work, we will further investigate the transferability of object information within the scene, and strive to better transform object information to assist scene recognition.

\bibliographystyle{IEEEtran.bst}
\bibliography{Mybib.bib}


\end{document}